\documentclass{article}
\usepackage{ijcai26}

\usepackage{times}
\usepackage{soul}
\usepackage{url}
\usepackage[hidelinks]{hyperref}
\usepackage[utf8]{inputenc}
\usepackage[small]{caption}
\usepackage{graphicx}
\usepackage{amsmath}
\usepackage{amsthm}
\usepackage{booktabs}
\usepackage{algorithm}
\usepackage{algorithmic}
\usepackage[switch]{lineno}

\usepackage{color}
\usepackage{multirow}
\usepackage{tcolorbox}
\usepackage{colortbl}
\usepackage{amsfonts}
\tcbuselibrary{breakable}

\title{EvoThink: Evolving Thinking in Large Reasoning Models via Self-Pruning and Aha-Moment Preference Optimization}

\author{
Xinbang Dai$^{1\ast}$\and
Zheyu Xin$^{1\ast}$\and
Huikang Hu$^{1\ast}$\and
Lin Ren$^{1}$\and
Rihui Jin$^{1}$\and \\
Guohui Xiao$^{1}$\and 
Guilin Qi$^{1}$\and
Kuicai Dong$^{2}$\and
Zhaocheng Du$^{2}$\and
Yuyang Zhang$^{2}$
\affiliations
$^1$Southeast University\\
$^2$Noah's Ark Lab
\emails
\{xbdai, huikanghu\}@seu.edu.cn
}

\begin{document}

\maketitle

\renewcommand{\thefootnote}{\fnsymbol{footnote}}
\footnotetext[1]{Equal contribution.}

\begin{abstract}
Large Reasoning Models (LRMs) often suffer from \textbf{overthinking} due to redundant verification steps. Existing approaches for mitigating overthinking, such as fast-slow thinking switching and reasoning trajectory compression, fail to make a fine-grained distinction between beneficial and redundant steps within the LRM's reasoning process, and may thus impair reasoning capability in their pursuit of efficiency. To simultaneously improve reasoning efficiency and capability, we propose \textsc{EvoThink}, a framework that reduces redundant verification and encourages the exploration of new reasoning paths. \textsc{EvoThink} comprises two key components: Self-Pruning Training (SPT), an unsupervised method that iteratively prunes redundant reasoning steps and self-trains on the concise trajectories; and Aha-Moment Preference Optimization (AMPO), which, inspired by genetic algorithms, identifies valuable failed reasoning attempts, synthesizes \emph{from-wrong-to-right} aha-moment data, and optimizes the model to internalize this reasoning pattern. Extensive evaluations across mathematical reasoning and code generation benchmarks demonstrate that \textsc{EvoThink} not only substantially reduces inference-time token usage but also improves the reasoning capability of LRMs.
\end{abstract}

\section{Introduction}
\begin{figure}[t]
  \centering
  \includegraphics[width=\columnwidth]{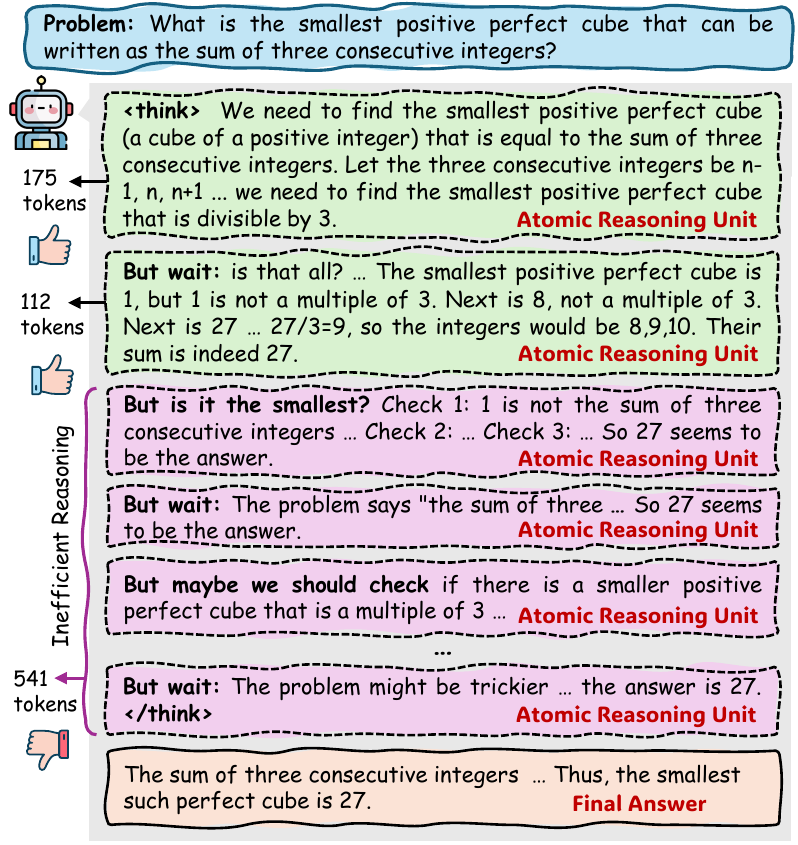}
  \caption{An example of overthinking in DeepSeek-R1. The model’s reasoning trajectory can be decomposed into a sequence of independent Atomic Reasoning Units: self-contained text segments, each capturing a single reasoning step and leading to a local conclusion. While beneficial units (green) effectively deepen the reasoning, inefficient units (purple) consist of redundant self-verification that consumes tokens without adding new insights.}
  \label{fig1}
\end{figure}

Large reasoning models (LRMs) have demonstrated significant performance improvements in solving complex tasks such as mathematical reasoning and code generation~\cite{guo2025deepseek,team2025kimi}. However, they often exhibit \textbf{overthinking}, wherein they repeatedly solve and verify the same problem during inference, leading to a substantial increase in output length~\cite{chen2024not,sui2025stop}. This verbose content incurs considerable computational costs and token consumption but offers minimal improvement in final task accuracy, sometimes even degrading performance through noise accumulation~\cite{cuadron2025danger}. As illustrated in Figure~\ref{fig1}, while some verification is beneficial for in-depth analysis, over 65\% of the tokens are spent on redundant computations that retrace previously established reasoning paths, substantially reducing inference efficiency.

Existing efforts to mitigate overthinking in LRMs fall into two broad categories. The first line of work focuses on switching between \emph{fast} and \emph{slow} thinking: for difficult problems, it aims to exploit the reasoning capability of LRMs via step-by-step inference, whereas for simple problems, it encourages the model to bypass reasoning and respond directly. In essence, this approach makes a binary decision before reasoning begins~\cite{pan2024dynathink,su2025dualformer}. However, it is inherently vulnerable to misclassification, which can propagate errors and lead to reasoning failures. The second line of work targets reasoning trajectory compression, aiming to reduce the number of tokens produced during inference~\cite{luo2025o1,hou2025thinkprune,chen2025the}. In particular, methods that leverage a model’s own capabilities to compress reasoning trajectories have attracted attention. They do not require supervision from human experts or stronger models; instead, they compress output length via lightweight heuristic rules during training. However, existing compression methods typically do not distinguish between beneficial verification steps that support deeper reasoning and redundant checks at a fine granularity. As a result, they tend to shorten reasoning trajectories indiscriminately, which may disrupt beneficial reasoning steps and thereby undermine the reasoning capability of LRMs. Overall, existing methods that switch reasoning modes based on problem difficulty or compress overall output length do not specifically address the issue of redundant reasoning steps, thereby limiting further improvements in reasoning efficiency. 

Moreover, prior efforts to mitigate overthinking largely overlook enhancing reasoning capability~\cite{cheng2025optimizing,hou2025thinkprune}. They typically regard the slight performance drop of the optimized model relative to a vanilla LRM--attributable to shorter reasoning traces--as an acceptable trade-off. This perspective suggests that current overthinking-mitigation strategies do not focus on recovering the model's potentially undermined reasoning ability.

To address these challenges, we propose \textsc{EvoThink}, a two-stage framework that optimizes both reasoning efficiency and capability. First, to improve reasoning efficiency, we introduce \textbf{Self-Pruning Training} (SPT), an unsupervised procedure that iteratively identifies and removes redundant reasoning steps and then performs self-training on the pruned trajectories. Second, we propose \textbf{Aha-Moment Preference Optimization} (AMPO) to enhance reasoning capability. Inspired by genetic algorithms, AMPO treats failed reasoning trajectories as a population and uses a fitness function to quantify the diversity of reasoning attempts, thereby selecting high-potential failures. Based on these curated failures, we construct aha-moment data that captures the \emph{from-wrong-to-right} transitions. We then apply direct preference optimization (DPO)~\cite{rafailov2023direct} to drive the model to internalize this reasoning pattern. We conduct extensive evaluations on mathematical reasoning and code generation benchmarks, revealing that \textsc{EvoThink} consistently improves both reasoning efficiency and capability. Furthermore, we find that, compared to optimizing the model using only correct solutions, our aha-moment data construction pattern better emulates the exploratory process of discovering new reasoning paths, thereby facilitating improvements in reasoning performance.
In summary, our contributions are:
\begin{itemize}
    \item We identify overthinking as a reasoning quality issue and introduce SPT, an unsupervised method to prune redundant verification steps from reasoning trajectories.
    \item We propose AMPO, a method to encourage exploration of new reasoning paths by learning from curated \emph{from-wrong-to-right} transitions on high-potential failures.
    \item Through extensive experiments on math and code benchmarks, we demonstrate that \textsc{EvoThink} significantly enhances both reasoning efficiency and overall task performance.
\end{itemize}
\section{Related Works}
\paragraph{Slow-fast thinking mode switching.}
This line of research typically assesses reasoning complexity to determine whether to invoke a fast or slow reasoning mode. Dynathink~\cite{pan2024dynathink} introduces a dynamic framework allowing an LLM to autonomously choose between strategies: a fast mode for tasks with quickly identifiable answers, and a slow mode for more complex problems. Dualformer~\cite{su2025dualformer} proposes controllable fast-slow reasoning by learning stochastic reasoning trajectories, enabling the model to adaptively select reasoning modes during inference. Similarly, ACPO~\cite{cheng2025incentivizing} introduces control tokens such as \texttt{<fast\_think>} and \texttt{<slow\_think>} to enable dynamic transitions between different reasoning depths. Thinkswitcher~\cite{liang2025thinkswitcher} determines when to engage in deep versus rapid reasoning by analyzing the impact of switching on computation time, thereby reducing decoding FLOPs. Despite their effectiveness, these approaches fundamentally rely on a classification mechanism to select an appropriate reasoning mode; this dependence may introduce binary decision errors and lead to selecting an incorrect reasoning mode due to error propagation.

\paragraph{Reasoning trajectory compression.}
Although complex reasoning is more likely to yield correct answers, the associated lengthy reasoning processes substantially increase both inference latency and computational costs. To promote efficient reasoning, recent studies incorporate efficiency considerations into training frameworks. For example, Kimi~\cite{team2025kimi}, O1-Pruner~\cite{luo2025o1}, and ThinkPrune~\cite{hou2025thinkprune} introduce length-based penalties into reinforcement learning reward functions, encouraging models to produce more concise reasoning while maintaining accuracy. In addition, DAST~\cite{shen2025dast} and DIET~\cite{chen2025the} construct preference datasets to train models to generate reasoning sequences that match the complexity of the query. Beyond training-based approaches, several training-free methods improve test-time reasoning efficiency via advanced prompting techniques, such as Sketch-of-Thought~\cite{aytes2025sketch}, Token-Budget~\cite{han2025token}, and No Think~\cite{ma2025reasoning}. However, these methods generally lack a fine-grained distinction between beneficial steps that deepen reasoning and redundant, unproductive steps. Consequently, compression-based strategies may indiscriminately remove essential information, potentially impairing the model's reasoning capability.
\begin{figure*}[t]
  \centering
  \includegraphics[width=0.95\textwidth]{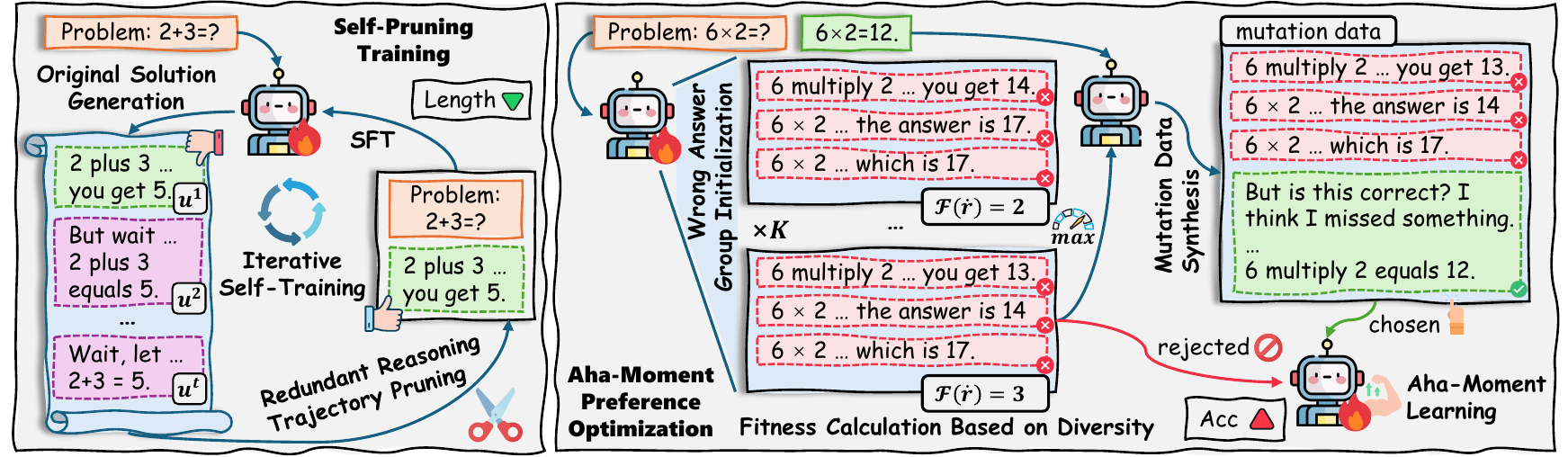}
  \caption{Self-Pruning Training learns concise reasoning by pruning redundant units, whereas Aha-Moment Preference Optimization uses a fitness function to select the most valuable failed reasoning trajectories, then synthesizes mutation examples and trains the model using DPO.}
  \label{fig2}
\end{figure*}

\section{Method}
As illustrated in Figure~\ref{fig2}, \textsc{EvoThink} comprises two sequential stages designed to improve the reasoning efficiency of LRMs by reducing redundant verification and optimizing reasoning patterns. The first stage, \textbf{Self-Pruning Training} (SPT), follows an iterative loop in which the model generates an initial solution, prunes redundant reasoning trajectories, and then performs self-training on the resulting pruned data. In the second stage, \textbf{Aha-Moment Preference Optimization} (AMPO), we initialize a population of incorrect responses and apply a fitness function $\mathcal{F}(.)$ to identify valuable failure cases. We then create mutated data by strategically connecting the incorrect reasoning paths in failure cases to the correct answers, yielding explicitly aha-moment instances. Finally, the model learns this reasoning pattern via DPO, thereby improving its reasoning capability.

\subsection{Self-Pruning Training}
\label{self_pruning_training}

\subsubsection{Original Solution Generation}
In this step, the LRM generates preliminary reasoning for a given query. Formally, we define $\mathcal{M}_{i}$ as the LRM at the $i$-th iteration of the self-training process. Specifically, $\mathcal{M}_{0}$ denotes the initial model prior to any optimization. This initial model serves as the annotator (denoted by $\mathcal{A}$) and processes all data using our task-specific prompts within the \textsc{EvoThink} framework. Given a query $q$, the model generates an output sequence formulated as $o_i = r_i \oplus \hat{a}_i$. Specifically,  $r_i$ represents the reasoning trajectory (enclosed within \texttt{<think>} and \texttt{</think>} tokens) capturing the model's step-by-step thought process, $\hat{a}_i$ represents the final predicted answer generated after the reasoning trajectory, and $\oplus$ signifies sequence concatenation.

\subsubsection{Redundant Reasoning Trajectory Pruning}
This step removes repetitive self-verification loops within the generated reasoning trajectory $r_i$ while preserving useful reasoning progress. We first introduce the concept of the \emph{atomic reasoning unit} to formalize the pruning process.

\paragraph{Atomic reasoning unit.}
Given $r_i$, we prompt $\mathcal{A}$ to decompose it into a sequence of complete reasoning sub-steps, denoted as $r_i = \{u_i^{(t)}\}$ for $t \in \{1, \dots, T\}$. Each $u_i^{(t)}$ constitutes an \emph{atomic reasoning unit}: a contiguous text span that starts when the model initiates a new self-contained reasoning attempt (understandable without relying on later units) and ends when that attempt reaches a local conclusion (e.g., a candidate direction, intermediate claim, or tentative answer). Empirically, with the exception of the first unit, such units often begin with discourse markers indicating a restart or re-check (e.g., \emph{wait}, \emph{alternatively}, \emph{but}, \emph{however}, \emph{double-check}).

\paragraph{Pruning redundant atomic units.}
The LRM first segments the reasoning trajectory into \emph{atomic reasoning units} and subsequently examines the redundancy of consecutive units. Concretely, the model compares the local conclusion implied by $u_i^{(t)}$ with that of $u_i^{(t-1)}$; if they are the same (i.e., the new unit does not expand the explored solution space and only re-derives or re-verifies an already reached conclusion), the model marks $u_i^{(t)}$ as redundant and removes it. This yields a pruned reasoning trajectory $\tilde{r}_i$, and the corresponding pruned output sequence $\tilde{o}_i = \tilde{r}_i \oplus \hat{a}_i$, where the predicted answer $\hat{a}_i$ remains unchanged during this step.

\paragraph{Pruning-induced efficient reasoning patterns.}
Our pruning strategy establishes an efficient reasoning pattern in the model. Specifically, (1) when a reasoning unit is consistently correct yet repeatedly revisited, pruning encourages the model to halt upon reaching the conclusion, thereby reducing unnecessary over-verification. (2) When a reasoning unit is incorrect yet repeatedly invoked, pruning removes consecutively repeated conclusions so that each attempt explores a new subspace, alleviating the model's over-concentration on high-likelihood but incorrect solutions. (3) When the model initially oscillates among incorrect conclusions but ultimately reaches the correct answer (i.e., an \emph{aha moment}), pruning reduces the number of prior repeated attempts spent on incorrect trajectories, enabling the reasoning process to reach the correct answer faster.

\subsubsection{Iterative Self-Training}
Following the $i$-th pruning step, the model $\mathcal{M}_{i-1}$ undergoes unsupervised fine-tuning on a set of concise solutions that it self-generates, yielding the subsequent model $\mathcal{M}_i$. The size of the concise solution set is fixed at 100 for each iteration. This design encourages the model to internalize the generalizable skill of concise reasoning (the \emph{process}) rather than simply memorize efficient paths to known correct answers (the \emph{results}). Let $\bar{L}_i$ denote the average token length of the pruned outputs generated by $\mathcal{M}_{i}$ across all queries at the $i$-th iteration. We define the relative length reduction as $S(i)=(\bar{L}_{i-1}-\bar{L}_i)/\bar{L}_{i-1}$ for $i \ge 1$. The procedure terminates when the compression gain becomes negligible, i.e., when $S(i) \le \epsilon_L$, where the threshold hyperparameter $\epsilon_L$ is set to 0.1. We justify this choice in \S~\ref{exp_reasoning_efficiency}.

The self-training process utilizes a standard next-token prediction loss, formulated as:
\begin{equation*}
\mathcal{L}_{\text{SPT}} = - \frac{1}{|\tilde{o}_i|} \sum_{w=1}^{|\tilde{o}_i|} \log P_{\mathcal{M}_{i}}(\tilde{o}_{i,w} \mid q, \tilde{o}_{i,<w})
\end{equation*}
where $q$ is the input query, $\tilde{o}_i$ denotes the pruned output sequence consisting of tokens $\tilde{o}_{i,w}$, and $P_{\mathcal{M}_{i}}$ represents the probability distribution of the model at iteration $i$. By minimizing this loss, the model learns to internalize a more efficient reasoning pattern.

\subsection{Aha-Moment Preference Optimization}
\label{aha_moment_preference_optimization}

\subsubsection{Wrong Answer Group Initialization}
In this stage, we employ a genetic algorithm to extract valuable learning signals from failures. These signals are then used to construct mutation data, which trains the LRM to acquire new reasoning patterns. The process starts with the model $\mathcal{M}_{I}$ obtained at the final iteration (the $I$-th round) of the SPT stage. We specifically select this warm-started model because it has internalized efficient reasoning patterns. This trait proves particularly advantageous when generating erroneous reasoning trajectories: compared to a vanilla model lacking such internalized patterns, $\mathcal{M}_{I}$ requires significantly fewer tokens to produce the same number of atomic reasoning units, as discussed in \S~\ref{exp_reasoning_efficiency}.

For each query $q$, we use this model to produce $K$ responses, where $K$ is a hyperparameter. If all $K$ attempts yield incorrect answers, we mark the query as a \emph{hard query}, representing the current reasoning boundary of the model. This process forms a subset $\mathcal{D}_{\text{hard}}$. For every $q \in \mathcal{D}_{\text{hard}}$, the $K$ incorrect reasoning outputs $\{\dot{o}_1, \dots, \dot{o}_k\}$ constitute the initial population for evolution, termed the \emph{Wrong Answer Group}. The choice of $K$ is detailed in \S~\ref{exp_size}.

\subsubsection{Fitness Calculation Based on Diversity}
To select the most informative failure cases for evolution, we define a diversity-based fitness function $\mathcal{F}$. For a reasoning trajectory $\dot{r}$, we first decompose it into $m$ atomic reasoning units, $\dot{r} = (u_1, u_2, \ldots, u_m)$. Let $\mathcal{A}$ be a parser that extracts from each unit $u_i$ a local conclusion $c_i = \mathcal{A}(u_i)$. We then define the set of distinct local conclusions in $\dot{r}$ as $C(\dot{r}) = \{c_i\}_{i=1}^{m}$, and compute the fitness as the number of unique conclusions: $\mathcal{F}(\dot{r}) = |C(\dot{r})| = \left|\{\mathcal{A}(u_i)\,:\, i=1,\ldots,m\}\right|$. A larger $\mathcal{F}(\dot{r})$ indicates that the trajectory explores a broader range of intermediate conclusions, and is therefore more informative. In practice, we prompt $\mathcal{A}$ to extract $\{c_i\}$ and count the number of distinct conclusions.

\subsubsection{Mutation Data Synthesis}
To construct valid mutation data, we select the erroneous trajectory with the highest fitness score and prompt the annotator $\mathcal{A}$ to synthesize an aha-moment transition. Specifically, for a given query $q$, the model retains the entire incorrect reasoning prefix, while we prompt it to initiate a shift toward the correct answer $a$ with the starting sentence, ``But is this correct? I think I missed something." This creates a synthetic trajectory that begins with plausible but flawed reasoning (aligning with the model's internal distribution) and gradually bridges to the correct solution (aligning with the gold answer distribution). Unlike direct supervision with correct answers, this \emph{from-wrong-to-right} pattern offers a softer learning curve, effectively mitigating the model's overconfidence in high-likelihood incorrect paths while demonstrating how to recover from errors. We analyze this setting in \S~\ref{exp_reasoning_paradigm_learning} and present a case study in \S~\ref{case_study}.

\subsubsection{Aha-Moment Learning}
To help the model learn the synthesized aha-moment reasoning patterns, we train it using DPO. For each hard query $q \in \mathcal{D}_{\text{hard}}$, we construct a preference pair $(o^{ch}, o^{re})$, where the \emph{chosen} response $o^{ch}$ is the synthesized mutation trajectory and the \emph{rejected} response $o^{re}$ is the original incorrect trajectory with the highest fitness score. This pairing forces the model to distinguish between a dead-end reasoning path and a successful aha-moment correction. For $q \notin \mathcal{D}_{\text{hard}}$, $o^{ch}$ is the shortest correct response among $K$ sampled outputs, while $o^{re}$ is the longest response, which is not necessarily correct. The objective of AMPO is formulated as:
{\small
\begin{equation*}
\mathcal{L}_{\text{AMPO}}= -\Biggl[ \log \sigma \Biggl( \beta \log \frac{P_{\mathcal{M}}(o^{ch} \mid q)}{P_{\text{ref}}(o^{ch} \mid q)} - \beta \log \frac{P_{\mathcal{M}}(o^{re} \mid q)}{P_{\text{ref}}(o^{re} \mid q)} \Biggr) \Biggr]
\end{equation*}
}
where $\mathcal{M}$ is the policy model under optimization (initialized from $\mathcal{M}_{I}$), and $P_{\text{ref}}$ is the reference model (typically a copy of $\mathcal{M}_{I}$ before DPO begins). The hyperparameter $\beta$ controls the strength of the preference optimization; we set $\beta$=0.5 to enforce closer fitting to the preference data. The function $\sigma(\cdot)$ is the standard sigmoid function. By minimizing $\mathcal{L}_{\text{AMPO}}$, the model is trained to assign higher probability to $o^{ch}$ relative to $o^{re}$.

\section{Experiments}

\begin{table*}[t]
\centering
\renewcommand{\arraystretch}{0.85}
\resizebox{0.88\textwidth}{!}{%
\begin{tabular}{llcccccccc}
\toprule
\multirow{2}{*}{\textbf{Backbone}} & \multirow{2}{*}{\textbf{Method}} &
\multicolumn{2}{c}{\textbf{MATH500}} &
\multicolumn{2}{c}{\textbf{AIME24}} &
\multicolumn{2}{c}{\textbf{AIME25}} &
\multicolumn{2}{c}{\textbf{TACO}} \\
\cmidrule(r){3-4}\cmidrule(r){5-6}\cmidrule(r){7-8}\cmidrule(r){9-10}
& &
\multicolumn{1}{c}{P@1 ↑} & \multicolumn{1}{c}{\#Tok ↓} &
\multicolumn{1}{c}{P@1 ↑} & \multicolumn{1}{c}{\#Tok ↓} &
\multicolumn{1}{c}{P@1 ↑} & \multicolumn{1}{c}{\#Tok ↓} &
\multicolumn{1}{c}{P@1 ↑} & \multicolumn{1}{c}{\#Tok ↓} \\
\midrule

\multirow{9}{*}{DeepScaleR-1.5B}
& Vanilla & 80.3 & 3171 & 29.2 & 6436 & 30.8 & 8282 & 4.2 & 7678 \\
& Kimi 1.5 SFT & 79.3 & 2911 & 27.5 & 6399 & 29.2 & 8185 & 3.5 & 7420 \\
& Kimi 1.5 DPO & 81.5 & 2883 & 35.0 & 6739 & 26.7 & 6751 & 3.7 & 7350 \\ 
& O1-Pruner & 79.6 & 2430 & 28.3 & 5205 & 33.3 & 6441 & 5.5 & 7622 \\
& ThinkPrune-2k & 80.1 & \textbf{1838} & 26.7 & \textbf{3862} & 27.5 & \textbf{4392} & 3.5 & \textbf{6117} \\
& DIET & 81.5 & 2255 & 29.2 & 5147 & 32.5 & 5881 & 5.8 & 7241 \\
\cmidrule(r){2-10}
& \textsc{EvoThink}\textsubscript{SPT} & 80.2 & \underline{1861} & 26.7 & \underline{4321} & 26.7 & \underline{5470} & 6.4 & \underline{6234} \\
& \textsc{EvoThink}\textsubscript{AMPO} & \textbf{84.7} & 3441 & \textbf{37.5} & 7044 & \textbf{37.5} & 8783 & \textbf{7.7} & 8013 \\
& \textsc{EvoThink}\textsubscript{SPT+AMPO} & \underline{82.1} & 2146 & \underline{36.7} & 5287 & \underline{35.8} & 6254 & \underline{7.4} & 6919 \\
\midrule

\multirow{9}{*}{Distill-Qwen-1.5B}
& Vanilla & 75.4 & 3752 & 16.7 & 7337 & 24.2 & 8319 & 4.2 & 7680\\  
& Kimi 1.5 SFT & 68.7 & 3643 & 17.5 & 6688 & 19.2 & 7120 & 3.4 & 7450 \\
& Kimi 1.5 DPO & 70.6 & 3688 & 20.0 & 7423 & 20.8 & 6976 & 3.6 & 7380 \\
& O1-Pruner & 69.7 & 3451 & 14.2 & 6436 & \underline{29.2} & 7371 & 4.1 & 7481 \\
& ThinkPrune-2k & 72.1 & \underline{2659} & 13.3 & \textbf{5402} & 20.8 & \textbf{5731} & 3.7 & 7029 \\
& DIET & 72.9 & 3142 & 16.7 & 6870 & 26.7 & 6702 & 3.5 & 7406 \\
\cmidrule(r){2-10}
& \textsc{EvoThink}\textsubscript{SPT} & 73.5 & \textbf{2481} & 20.0 & \underline{5988} & 22.5 & \underline{5946} & 5.2 & \textbf{6645} \\
& \textsc{EvoThink}\textsubscript{AMPO} & \textbf{78.0} & 3827 & \textbf{35.8} & 8071 & \textbf{30.0} & 9151 & \textbf{6.6} & 8448 \\
& \textsc{EvoThink}\textsubscript{SPT+AMPO} & \underline{76.3} & 2778 & \underline{30.8} & 6633 & 28.3 & 6572 & \underline{6.2} & \underline{7024} \\
\midrule

\multirow{9}{*}{QwQ-32B}
& Vanilla & 87.6 & 4289 & 46.7 & 7290 & 33.3 & 8749 & 10.7 & 9003 \\
& Kimi 1.5 SFT & 87.4 & 4115 & 44.2 & 7025 & 31.7 & 7409 & 10.5 & 8735 \\
& Kimi 1.5 DPO & 85.8 & 4075 & 42.5 & 6950 & 30.8 & 7330 & 10.0 & 8640 \\
& O1-Pruner & 83.1 & 3002 & 37.5 & 6939 & 27.5 & 6598 & 8.4 & 8472 \\
& ThinkPrune-2k & 88.7 & \textbf{2336} & 45.8 & \textbf{5824} & 32.5 & 6465 & 8.7 & \underline{7198} \\
& DIET & 83.6 & 3740 & 49.2 & 6314 & 34.2 & 7125 & 9.8 & 7219 \\
\cmidrule(r){2-10}
& \textsc{EvoThink}\textsubscript{SPT} & 86.0 & \underline{2406} & 50.8 & \underline{6225} & 35.8 & \textbf{6224} & 9.4 & \textbf{6947} \\
& \textsc{EvoThink}\textsubscript{AMPO} & \textbf{90.1} & 4718 & \textbf{55.8} & 8019 & \textbf{38.3} & 9224 & \textbf{12.6} & 9903 \\
& \textsc{EvoThink}\textsubscript{SPT+AMPO} & \underline{89.5} & 3016 & \underline{52.5} & 6299 & \underline{36.7} & \underline{6395} & \underline{11.9} & 7281 \\
\bottomrule
\end{tabular}
}
\caption{Average performance (Pass@1, \%) and token length (\#Tok) across different models and baselines on four benchmarks. For each benchmark, \textbf{bold} denotes the best result and \underline{underline} denotes the second-best.}
\label{tab1}
\end{table*}

\subsection{Experimental Setups}
\paragraph{Baselines.}
In our experiments, we evaluate three open-source backbone LRMs: Distill-Qwen-1.5B~\cite{guo2025deepseek}, DeepScaleR-1.5B~\cite{deepscaler2025}, and QwQ-32B~\cite{yang2024qwen}. We compare \textsc{EvoThink} with three groups of baselines. (1) The vanilla model without compression. (2) Supervised Fine-tuning (SFT) and DPO baselines: Kimi 1.5 SFT\cite{team2025kimi}, fine-tuned on the shortest correct responses, and Kimi 1.5 DPO\cite{team2025kimi}, which treats the shortest correct response as a positive example and incorrect or longer ones as negative examples. (3) RL-based methods, including O1-Pruner~\cite{luo2025o1}, ThinkPrune~\cite{hou2025thinkprune}, and DIET~\cite{chen2025the}. We follow the settings established in these baselines. For ThinkPrune, we adopt the ThinkPrune-2k variant from the series, which achieves the greatest reduction in output length.

\paragraph{Datasets.}
We comprehensively evaluate all baselines on four benchmarks covering distinct domains: MATH-500~\cite{hendrycks2021measuring},  AIME24~\cite{muennighoff2025s1} and AIME25\footnote{\url{https://huggingface.co/datasets/math-ai/aime25}} (mathematics), as well as TACO~\cite{li2023taco} (code generation). For MATH-500, we train exclusively on the standard MATH training set. For AIME24 and AIME25, we construct a training set from AIME (1983–2023).  We obtain the gold solutions for all AIME problems from the Art of Problem Solving community\footnote{\url{https://artofproblemsolving.com/wiki/index.php?title=AIME\_Problems\_and\_Solutions}}. For TACO, we use its official training sets. To ensure a fair comparison, all baselines are trained and evaluated on identical data splits. We use veRL~\cite{sheng2025hybridflow} as the training framework and train models on 8 H100 GPUs.

\paragraph{Evaluation.}
Following the established metrics from DIET~\cite{chen2025the}, we primarily evaluate performance using two metrics: Pass@1 (P@1) and the average response length in tokens (\#Tok). To estimate P@1, we sample 8 independent responses per question for AIME and 4 for all other datasets. For MATH-500 and AIME, we require the model to enclose its final answer in \texttt{\textbackslash \textbackslash boxed\{\}} to enable exact matching with the ground-truth answer; for TACO, we use the official evaluation script provided by the dataset. During the evaluation, we employed a temperature of 0.6, a top-p value of 0.95. For both the baseline and \textsc{EvoThink}, we set the maximum token limit per instance to 32,768 tokens. If generation reaches this limit before completing the reasoning process, we mark the instance as incorrect and set its response length to 32,768. This policy ensures that models failing to produce a complete solution within the token limit are penalized during accuracy computation.

\subsection{Main Results}
Table~\ref{tab1} presents a comprehensive quantitative evaluation of our \textsc{EvoThink} framework across diverse LRMs and benchmarks. The integrated method, \textsc{EvoThink}\textsubscript{SPT+AMPO}, consistently achieves a superior balance between reasoning accuracy (Pass@1) and computational efficiency (token length). When comparing fine-grained performance, we observe that while baselines like ThinkPrune-2k achieve substantial token reduction (e.g., reducing DeepScaleR-1.5B to 1838 tokens on MATH500), they often suffer significant performance degradation. For instance, on the MATH-500, when the vanilla model is DeepScaleR-1.5B, ThinkPrune-2k's Pass@1 drops to 72.1\% compared to the vanilla model's 75.4\%, whereas our \textsc{EvoThink}\textsubscript{SPT+AMPO} maintains a higher Pass@1 of 76.3\% with comparable efficiency gains. Overall, our approach outperforms other compression-focused baselines in Pass@1 while maintaining similar output lengths, demonstrating consistency across varying model scales.

The ablation variants, \textsc{EvoThink}\textsubscript{SPT} and \textsc{EvoThink}\textsubscript{AMPO}, represent models trained exclusively with either SPT or AMPO, respectively. Through the \textsc{EvoThink}\textsubscript{SPT} stage alone, the LRM achieves reasoning performance comparable to or even better than most existing baselines while effectively reducing token consumption. For instance, on the TACO dataset, our DeepScaleR-1.5B model attains a score of 6.4\% during the \textsc{EvoThink}\textsubscript{SPT} stage, outperforming all baselines. More importantly, \textsc{EvoThink}\textsubscript{SPT} training is unsupervised, requires only a few iterations, and uses much less data, whereas all competing baselines rely on gold answers and are trained on the full dataset. This suggests that, in addressing overthinking, improving reasoning quality by suppressing redundant atomic reasoning units is a more effective strategy than merely compressing the length of reasoning trajectories. 
Furthermore, \textsc{EvoThink}\textsubscript{AMPO} yields substantial performance gains; for example, on AIME24, QwQ-32B achieves 55.8\% Pass@1, compared to 46.7\% for the vanilla model. Consequently, the combined \textsc{EvoThink}\textsubscript{SPT+AMPO} framework achieves optimal overall performance, confirming the complementary advantages of its components. SPT minimizes token consumption, while AMPO significantly boosts the performance of the warm-started efficient models, thereby achieving a robust balance between performance and efficiency.

\begin{figure}[htbp]
  \centering
  \includegraphics[width=\columnwidth]{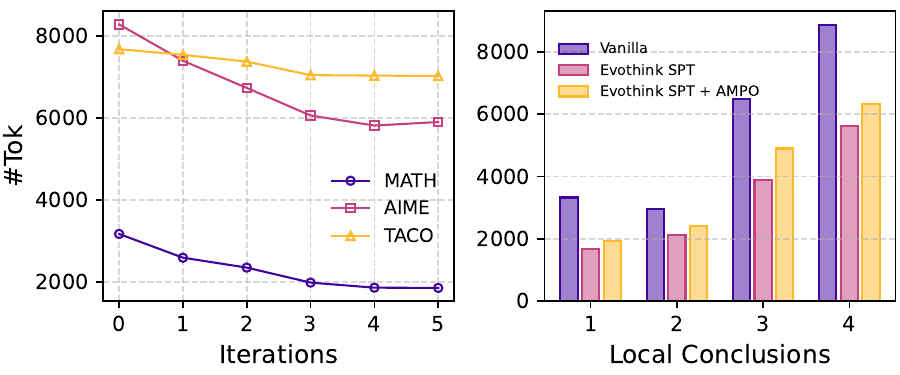}
  \caption{Evolution of reasoning length under \textsc{EvoThink}\textsubscript{SPT} on MATH (left). Correlation between reasoning length and the number of local conclusions on MATH (right).}
  \label{fig3}
\end{figure}

\subsection{Improvement of Reasoning Efficiency}
\label{exp_reasoning_efficiency}

Figure~\ref{fig3} (left) presents the evolution of the generated reasoning length of DeepScaleR-1.5B on the MATH-500 training set under the SPT algorithm. We observe that the length typically stabilizes within five iterations, at which point $S(i) < 0.1$. Accordingly, we set the length-tolerance threshold $\epsilon_L = 0.1$. On other datasets, the number of iterations required to stabilize and satisfy $S(i) < 0.1$ remains within ten. Figure~\ref{fig3} (right) reports the average reasoning length associated with local conclusions produced by DeepScaleR-1.5B on the MATH training set under different methods. For the vanilla model, no strict correlation exists between the number of local conclusions and the inference length. Notably, the token cost for a single local conclusion can even exceed that for two local conclusions in the vanilla model, which indicates redundant atomic reasoning units. After applying SPT, our framework captures the relationship between local conclusions and reasoning token cost: the reasoning length increases monotonically as the number of local conclusions grows.

\begin{figure}[htbp]
  \centering
  \includegraphics[width=0.9\columnwidth]{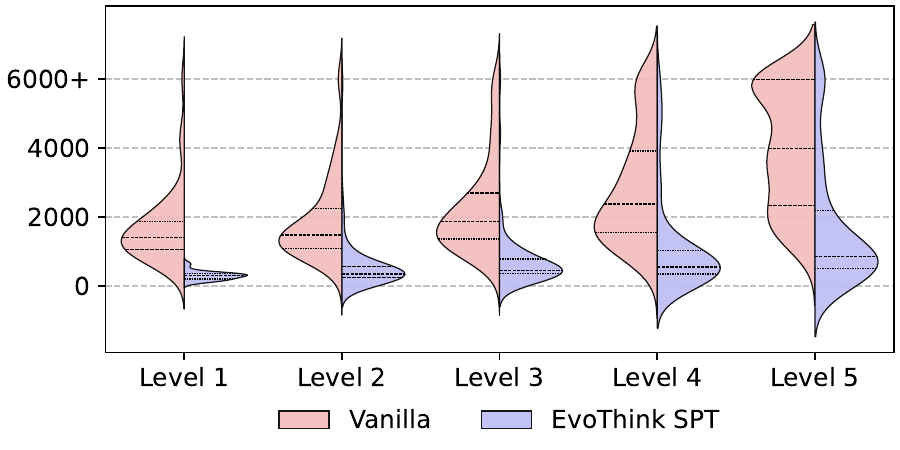}
  \caption{Distribution of reasoning lengths across difficulty levels on MATH-500.}
  \label{fig4}
\end{figure}

Figure~\ref{fig4} analyzes the output length of DeepScaleR-1.5B on MATH-500, grouped by problem difficulty (Levels 1–5, from easy to difficult). We find that: (1) \textsc{EvoThink}\textsubscript{SPT} yields substantially lower medians and interquartile ranges than the vanilla model at every difficulty level, indicating an effective reduction in reasoning length. (2) as difficulty increases, the vanilla model exhibits an increasingly heavy upper tail, suggesting that harder instances more frequently trigger extreme computational overhead. In contrast, \textsc{EvoThink}\textsubscript{SPT} shows a much milder increase with smaller variance; although its distribution shifts upward with difficulty, the magnitude is far smaller than that of the vanilla model. Overall, \textsc{EvoThink}\textsubscript{SPT} remains more stable and is less prone to excessive output length inflation.

\subsection{Enhancement of Reasoning Capability}
\label{exp_reasoning_paradigm_learning}

\begin{figure}[htbp]
  \centering
  \includegraphics[width=\columnwidth]{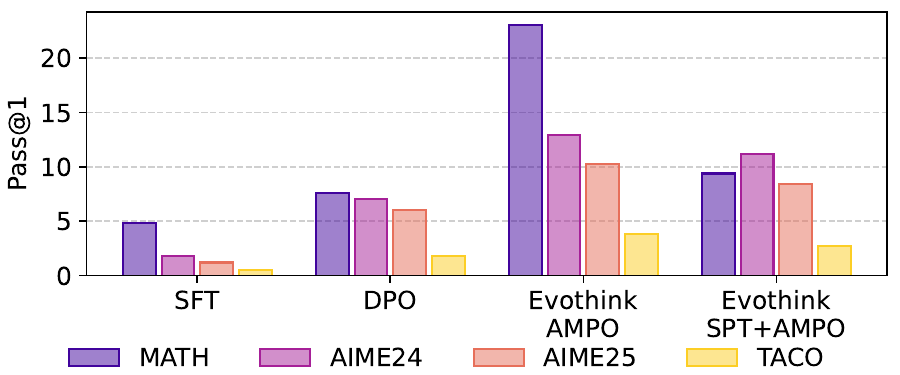}
  \caption{Performance improvements of DeepScaleR-1.5B under various training methods on the set of problems initially unanswered by the vanilla model.}
  \label{fig5}
\end{figure}

Figure~\ref{fig5} compares different training methods on curated difficult datasets composed of problems that the vanilla DeepScaleR-1.5B model consistently fails to solve. This setup allows for a focused analysis of how reasoning capabilities are improved by learning from gold solutions. We compare four approaches: (1) Supervised Fine-Tuning (SFT) directly on gold solutions; (2) DPO, using gold solutions as \emph{chosen} and model generations as \emph{rejected} responses; and (3, 4) our \textsc{EvoThink}\textsubscript{AMPO} and \textsc{EvoThink}\textsubscript{SPT+AMPO} methods, which leverage our constructed mutation data. The results show that, although training directly on human-written, highly condensed gold solutions substantially reduces reasoning length, SFT and DPO yield only limited improvements in reasoning capability, indicating that the model struggles to internalize this reasoning pattern through direct supervision. In contrast, the AMPO framework proves highly effective: our curated aha-moment data—enriched with diverse reasoning trajectories and aligned with gold solutions—substantially outperforms direct training on gold solutions alone.

To disentangle whether the observed improvement in reasoning stems from either memorizing answers or optimizing the reasoning pattern, we train DeepScaleR-1.5B exclusively on the MATH dataset and evaluate it on AIME24 and AIME25. As reported in Table~\ref{tab2}, after adopting the transfer learning setting, all baselines exhibit performance degradation to varying degrees. In contrast, \textsc{EvoThink}\textsubscript{AMPO} achieves a performance gain, further indicating that our method produces a more generalizable model. We attribute this improvement to high-quality synthetic data generated by our genetic algorithm, which teaches the model a robust reasoning strategy rather than rote memorization of solutions. Moreover, when combined with the SPT stage, \textsc{EvoThink}\textsubscript{SPT+AMPO} reduces output length while preserving its generalization capability.
We further find that training on human-written gold solutions effectively reduces token cost; however, the reasoning style in these solutions differs substantially from the model’s native reasoning patterns, and this mismatch markedly undermines LRM’s generalization. As shown in Table~\ref{tab2}, when models are trained on MATH with gold solutions, DPO reduces the Pass@1 score on the AIME to less than one-fifth of that of the untrained vanilla model.

\begin{table}[htbp]
\renewcommand{\arraystretch}{0.85}
\resizebox{0.9\columnwidth}{!}{%
\begin{tabular}{lcccc}
\toprule
 & \multicolumn{2}{c}{\textbf{AIME24}} & \multicolumn{2}{c}{\textbf{AIME25}} \\
 \cmidrule(r){2-3} \cmidrule(r){4-5}
 & P@1 ↑ & \#Tok ↓& P@1 ↑& \#Tok ↓\\
 \midrule
Vanilla & 29.2 & 6436 & 30.8 & 8282 \\
SFT & 4.2 & \underline{4965} & 3.3 & \textbf{6414} \\
DPO & 5.4 & 5342 & 4.6 & \underline{6778} \\
O1-Pruner & 17.5 & 5937 & 16.3 & 7451 \\
ThinkPrune-2k & 23.8 & 5542 & 24.2 & 6988 \\
DIET & 26.3 & 5814 & 25.0 & 7861 \\ 
\midrule
\textsc{EvoThink}\textsubscript{SPT} & 28.3 & \textbf{4811} & 30.4 & 7022 \\
\textsc{EvoThink}\textsubscript{AMPO} & \textbf{30.0} & 6492 & \textbf{31.3} & 8330 \\
\textsc{EvoThink}\textsubscript{SPT+AMPO} & 28.8 & 5005 & 30.8 & 7496 \\
\bottomrule
\end{tabular}
}
\caption{Transfer learning results on AIME24 and AIME25.}
\label{tab2}
\end{table}

\subsection{Analysis of Mutation Data}
\label{exp_size}

\begin{figure}[htbp]
  \centering
  \includegraphics[width=\columnwidth]{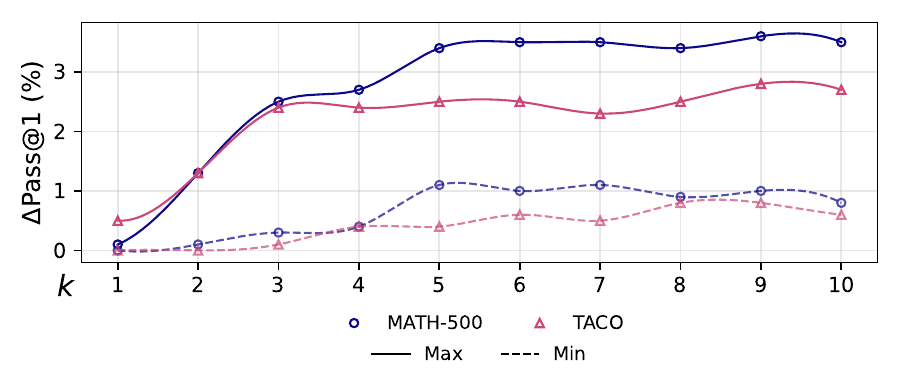}
  \caption{Performance gain vs. initial population size and reasoning trajectory diversity.}
  \label{fig6}
\end{figure}

Figure~\ref{fig6} illustrates the impact of the initial population size ($K$) on the final performance of \textsc{EvoThink}\textsubscript{SPT+AMPO} with the QwQ-32B model, as evaluated on the MATH-500 and TACO datasets. We observe that as $K$ increases, the variation in $\Delta$Pass@1 tends to stabilize for $K>6$. This observation is consistent across other models and datasets in our experiments; therefore, we set $K=6$. 

Furthermore, Figure~\ref{fig6} compares the performance of selecting the reasoning trajectory with maximum diversity (Max, i.e., $\max(\mathcal{F}(r_k))$ for $k \in {1, \dots, K}$) with that of selecting the one with minimum diversity (Min, i.e., $\min(\mathcal{F}(r_k))$). The results clearly show that high-diversity trajectories yield substantially larger performance gains than their low-diversity counterparts. This finding confirms that high-diversity trajectories are more suitable for use as mutation data, underscoring the importance of actively exploring new solution spaces to enhance the reasoning capabilities of LRMs.

\subsection{Case Study}
\label{case_study}

\begin{figure}[htbp]
  \centering
  \includegraphics[width=0.85\columnwidth]{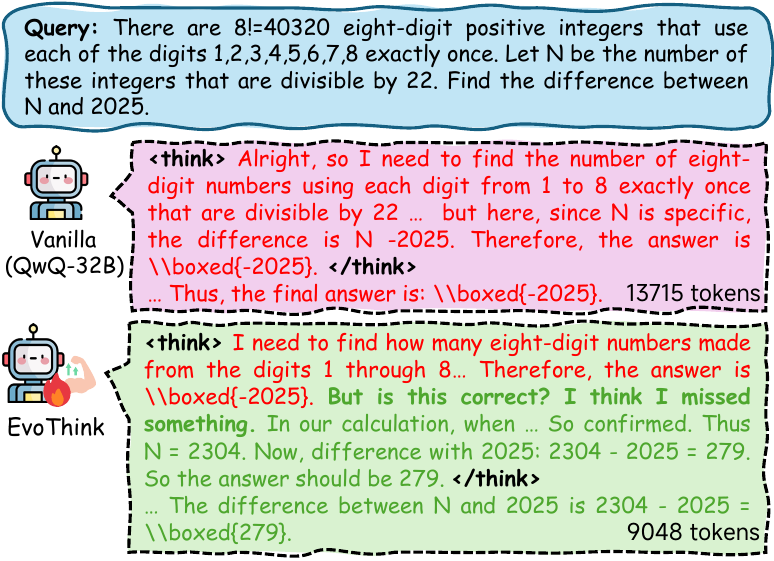}
  \caption{A case study on a math problem from AIME25.}
  \label{fig7}
\end{figure}

Figure~\ref{fig7} presents a case study of the QwQ-32B model solving an AIME25 problem. The vanilla model fails to execute the necessary reasoning steps, resulting in an incorrect answer and a lengthy output of 13,715 tokens. In contrast, \textsc{EvoThink} demonstrates the ability to explore new reasoning paths. As revealed in its thought process, it initially falls into a similar trap but then questions its own conclusion: ``But is this correct? I think I missed something." This moment enables it to discard the erroneous path and perform the correct calculations, requiring only 9,048 tokens—a 34\% reduction compared to the vanilla model.

\section{Conclusion and Future Work}
We introduce \textsc{EvoThink}, a framework that significantly enhances the reasoning efficiency and capability of LRM through SPT and AMPO. Our approach leverages a \emph{from-wrong-to-right} data construction process, allowing models to learn from failures and discover new reasoning paths. Extensive evaluations across math and code generation benchmarks demonstrate that \textsc{EvoThink} preserves performance, substantially reduces the average token count, and improves reasoning capability. For future work, we plan to investigate why the \emph{from-wrong-to-right} learning pattern adopted by AMPO is particularly effective for model training. This exploration is outside the scope of this paper but represents a vital direction for LRM reasoning.

\section*{Acknowledgements}
This work is partially supported by National Nature Science Foundation of China under No. 62476058. We thank the Big Data Computing Center of Southeast University for providing the facility support on the numerical calculations in this paper.

\bibliographystyle{named}
\bibliography{ref}

\end{document}